\documentclass{article}

\PassOptionsToPackage{numbers, compress}{natbib}

\usepackage[preprint]{neurips_2019}

\usepackage[utf8]{inputenc} 
\usepackage[T1]{fontenc}    
\usepackage{hyperref}       
\usepackage{url}            
\usepackage{booktabs}       
\usepackage{amsfonts}       
\usepackage{nicefrac}       
\usepackage{microtype}      
\usepackage{amsmath}
\usepackage{dblfloatfix}
\usepackage{xcolor}
\usepackage{booktabs}
\usepackage{epsfig}
\usepackage{graphicx}
\usepackage{kotex} 

\newcommand*{\affaddr}[1]{#1} 

\newcommand*{\email}[1]{\texttt{#1}}

\newcommand*\samethanks[1][\value{footnote}]{\footnotemark[#1]}

\title{Unpaired Image Translation via Adaptive Convolution-based Normalization}

\author{
    Wonwoong Cho\thanks{These authors contributed equally}, 
    Kangyeol Kim\samethanks, 
    Eungyeup Kim, 
    Hyunwoo J. Kim,
    Jaegul Choo \\
    \affaddr{Korea University}\\
    \small{\email{\{tyflehd21,kky1994,yhy1254,hyunwoojkim,jchoo\}@korea.ac.kr}}\\
}

\begin{document}

\maketitle

\begin{abstract}
Disentangling content and style information of an image has played an important role in recent success in image translation. In this setting, how to inject given style into an input image containing its own content is an important issue, but existing methods followed relatively simple approaches, leaving room for improvement especially when incorporating significant style changes. In response, we propose an advanced normalization technique based on adaptive convolution (AdaCoN), in order to properly impose style information into the content of an input image. In detail, after locally standardizing the content representation in a channel-wise manner, AdaCoN performs adaptive convolution where the convolution filter weights are dynamically estimated using the encoded style representation. The flexibility of AdaCoN can handle complicated image translation tasks involving significant style changes. 
Our qualitative and quantitative experiments demonstrate the superiority of our proposed method against various existing approaches that inject the style into the content. 

\end{abstract}
\section{Introduction}\label{sec:introduction}
Recently, unpaired image-to-image translation~\cite{Zhu_2017,kim2017learning,StarGAN2018} has been actively studied as one of the major research areas. It aims to learn inter-domain mappings without paired images, such that deep neural networks can translate a given image from one domain to another (e.g., real photo ${\Rightarrow}$ artwork). However, these methods bear a fundamental limitation of generating a uni-modal output given a single image even if multiple diverse outputs may exist. In response, several approaches~\cite{lin2018conditional,Huang_2018_ECCV,Lee_2018_ECCV,Xiao_2018_ECCV,ma2018exemplar,Chang:2018:PAS} have been proposed to achieve the multi-modality that indicates the capability of generating multiple outputs given a single input image by taking an additional input, such as an exemplar image conveying detailed style information to transfer.

Although exemplar-based image translation achieves multi-modality of outputs owing to its flexibility in reflecting the exemplar image that gives fine details of intended style, there still remains the issue of how to properly impose the style feature extracted from an exemplar image into a content image. Previous approaches~\cite{Lee_2018_ECCV,huang2017arbitrary,cho2019image} commonly follows two steps of first standardizing features and then applying a particular transformation, where the first step can be regarded as removing the existing style information of an input image
and the second step plays a role of imposing the exemplar style to the style-neutralized input feature.



As one of the state-of-the-art methods, adaptive instance normalization (AdaIN)~\cite{huang2017arbitrary} has been successfully utilized to combine content and style in a slew of studies~\cite{Huang_2018_ECCV,ma2018exemplar,karras2018style}. AdaIN incorporates different features by matching each channel's first-order statistics , e.g., the mean and the variance, in the content to those in the style. To this end, AdaIN first standardizes each channel of content feature and adaptively performs channel-wise scaling and shifting using the parameters regressed by the style feature. 
Another recently proposed method called group-wise deep whitening-and-coloring transformation (GDWCT)~\cite{cho2019image} has shown superior capability of imposing drastically different styles by matching higher-order statistics such as covariance, which we call the coloring transformation, in addition to the first-order ones. 

\begin{figure}[t]
  \includegraphics[width=\linewidth]{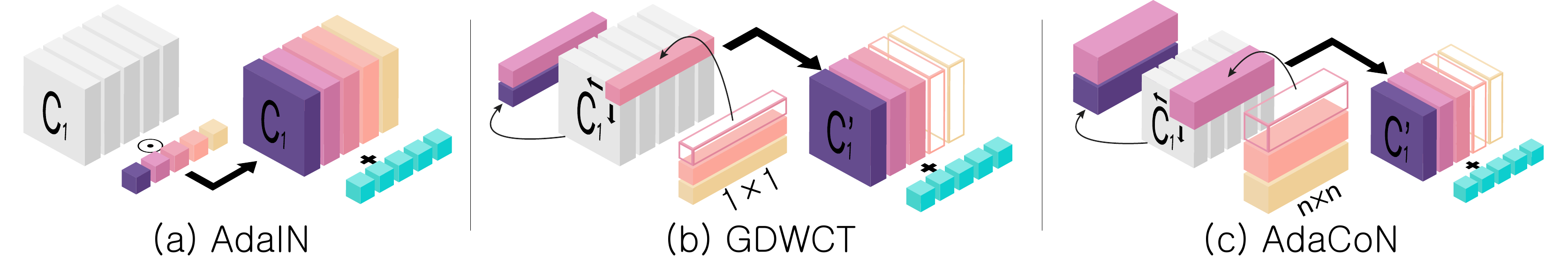}
  \vspace*{-0.5cm}
  \caption{Comparisons of different normalization methods for image translation. Each existing method can be viewed as the special case or the variant of a convolution operation. 
  }
  \label{Fig:Baseline comparison}
  \vspace*{-0.4cm}
\end{figure}

In the above methods, we claim that the second step of imposing the target statistics can be viewed as a simpler variant or a special case  of
a convolution operation, as illustrated in Fig.~\ref{Fig:Baseline comparison}. That is, (a) the channel-wise affine transformation used in AdaIN can be viewed as the channel-wise $1\times 1$ convolution. 
(b) On the other hand, the coloring transformation of GDWCT, which matches the target covariance, can be considered as the
$1 \times 1$ convolution operation that generates each output channel as a linear combination of the entire input channels.\footnote{The additional illustration can be found in Appendix.}
However, these methods tend to fail in handling a dramatic shape change because the methods have limited capability in translating significant transfiguration.

From this unified perspective of a convolution operation, these existing methods relied only on its simpler forms with only using $1\times 1$ convolution filters, and thus, the potentials of leveraging general convolution operations with larger-than-$1\times 1$ filters when injecting target style has not yet been fully explored. 

Inspired by this, we propose an adaptive convolution-based normalization (AdaCoN) as an advanced method to inject the target style to a given image. AdaCoN is basically composed of two steps of standardization and adaptive convolution. First, the standardization is locally performed on each sub-region of an input activation map where the convolution filter is applied, similar to previous work~\cite{jarrett2009best,krizhevsky2012imagenet}. Second, AdaCoN performs adaptive convolution where the (larger-than-$1\times 1$) convolution filter weights are dynamically estimated using the encoded style representation.

By taking into account spatial patterns due to a convolution operation, we hypothesize that AdaCoN is capable of flexibly performing a spatially-adaptive image translation, which can potentially handle complicated image translation tasks involving significant style changes. In this sense, AdaCoN has something in common with the recent success in patch-based style transfer~\cite{chen2016fast, gu2018arbitrary} that dynamically applies different styles to each patch of an input image. 

In order to verify the superiority of AdaCoN, we conduct both quantitative and qualitative experiments that compare different normalization methods while maintaining the same model architectures. 

\section{Related work}

\paragraph{Unpaired image translation.}
 Unpaired image translation aims to transform an input image from one domain to another without paired images. Numerous approaches~\cite{Zhu_2017,kim2017learning,liu2017unsupervised} have been proposed for this task. Recently, multimodal image translation methods, capable of yielding multiple different images given a particular image, have also been studied~\cite{Huang_2018_ECCV,Lee_2018_ECCV,cho2019image}. These studies take similar approaches to address the uni-modality problem of previous methods by incorporating an exemplar image as a guidance for image translation. In addition, they assume that a latent image space can be disentangled into the content space that contains an underlying structure of images and the style space that maintains a domain-specific feature. However, they propose different methods for integrating the disentangled content feature from the input image and the style feature from the exemplar image. To be specific, inspired by AdaIN~\cite{huang2017arbitrary}, MUNIT~\cite{Huang_2018_ECCV} adopts the idea of matching the statistics between the content and the style features. Extending this idea, GDWCT~\cite{cho2019image} leverages higher-order statistics compared to the previous method, enhancing the quality of generated images. Meanwhile, DRIT~\cite{Lee_2018_ECCV} simply concatenates the content and the style features to perform image translation. However, these methods have a limited capability to handle the drastic changes between the domains. A recently proposed method called instaGAN~\cite{mo2018instagan} tackles this problem by taking the segmentation mask as additional input, which serves as strong hint for translation. 

\paragraph{Adaptive convolution.} 
Unlike standard convolution layers where the filter weights are trainable constant values, an adaptive convolution layer uses varying filter weights dynamically determined by input data. Based on this idea, dynamic filter networks~\cite{jia2016dynamic} proposed to take an auxiliary input image to determine convolution filter weights in an video prediction task. Furthermore, Kang et al.~\cite{kang2017incorporating} showed that   convolution filter weights from the side information such as camera perspective or noise level can be utilized to improve the performance of classification task. Recent studies proposed to apply adaptive convolution to a variety of tasks such as semantic segmentation~\cite{harley2017segmentation,su2019pixel} and motion prediction~\cite{xue2016visual}. In this paper, we propose AdaCoN, which adaptively obtains convolution weights associated with convolution-based normalization for an image translation task.

\begin{figure}[t]
  \includegraphics[width=\linewidth]{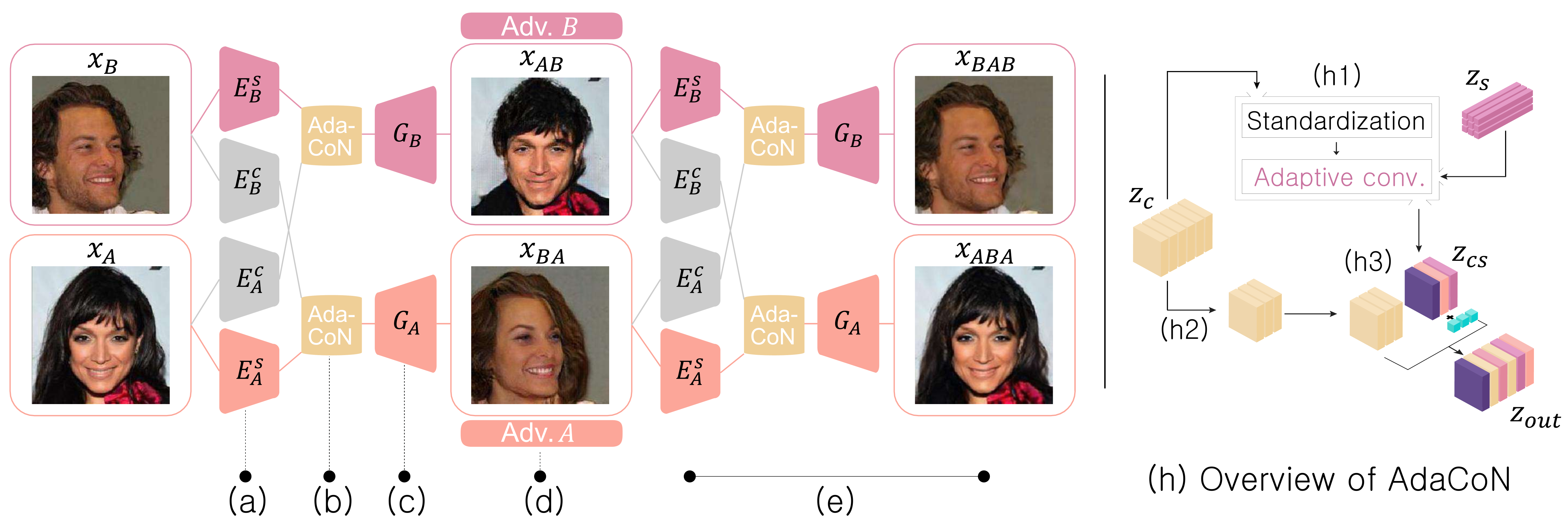}
  \vspace*{-0.6cm}
  \caption{Overview of our networks. 
  }
  \label{Fig:Overview of the networks}
  \vspace*{-0.4cm}
\end{figure}

\section{Proposed Methods}\label{sec:Methods}
In this section, we briefly describe our backbone networks for an image translation task. Afterwards, we concretely describe our proposed method in detail. 
\subsection{Translation backbone}\label{sub:Translation backbone}
\paragraph{Networks overview.}
Let ${x_{A}}$ and ${x_{B}}$ denote randomly sampled images from two different domains of ${\mathcal{X}_A}$ and ${\mathcal{X}_B}$, respectively. Given two images, our networks translate ${x_{A}}$ from domain ${\mathcal{X}_A}$ to domain ${\mathcal{X}_B}$ as well as ${x_{B}}$ from domain ${\mathcal{X}_B}$ to domain ${\mathcal{X}_A}$. 
To this end, we adopt the disentangling strategy~\cite{Huang_2018_ECCV,Lee_2018_ECCV,cho2019image} that decomposes an image into a domain-invariant content feature (e.g., an identity of a person) and a domain-specific style feature (e.g., the hair length in the female domain). This can be formulated as
\begin{align}
    z^c_A, z^s_A = E_A^c(x_A), E_A^s(x_A), \qquad\qquad z^c_B, z^s_B = E_B^c(x_B), E_B^s(x_B),
\end{align}
where \{${E_A^c}$ ,${E_B^c}$\} are content encoders and \{${E_A^s}$, ${E_B^s}$\} are style encoders. By combining the content and the style features of the different domains \{${(z^c_B,z^s_A)}$, ${(z^c_A,z^s_B)}$\} and forwarding it to decoders \{${G_A, G_B}$\}, we obtain the translated results \{${x_{B\rightarrow A}}$, ${x_{A\rightarrow B}}$\}, i.e.,
\begin{align}
x_{A\rightarrow B}=G_B(\textrm{AdaCoN}(z^c_A,z^s_B)),\qquad\qquad x_{B\rightarrow A}=G_A(\textrm{AdaCoN}(z^c_B,z^s_A)),
\end{align}
where ${\textrm{AdaCoN}}$ indicates our adaptive convolution-based normalization that incorporates given content and style features. As shown in Fig.~\ref{Fig:Overview of the networks}, for example, given ${\mathcal{X}_{A}}$ and ${\mathcal{X}_{B}}$ in the woman and the man domains, respectively, let us assume that our networks translate the woman to the man ${x_{A\rightarrow B}}$. (a) We first extract the content feature ${z^c_A}$ from a woman image ${x_A}$ and the style feature ${z^s_{B}}$ from a man image ${x_B}$ by forwarding each image into the content encoder ${E_A^c}$ and the style encoder ${E_B^s}$. (b) We next inject the style to the content feature through AdaCoN and (c) forward the combined features into the decoder ${G_A}$. After obtaining a fake man image ${x_{AB}}$, (d) we exploit the fake image as an input of a discriminator ${D_A}$ that encourages the generated image distribution to be close to the real image distribution. (e) Lastly, we repeat the processes of (a)-(c) in order to obtain a reconstructed woman image ${x_{ABA}}$, enabling our networks to maintain an original identity. In this manner, our networks are trained to translate the images between two different domains. 

\paragraph{Loss functions.}
Our networks are composed of several losses, and each term plays a crucial role in appropriately training our networks. In order to avoid redundancy, we focus on a translation of ${(\mathcal{X}_A\rightarrow{\mathcal{X}_B}\rightarrow{\mathcal{X}_A})}$ from this point on. First, we leverage the pixel-level reconstruction losses, such as the cycle-consistency loss and the identity loss~\cite{Zhu_2017} in order to guarantee the high-quality of generated images. The image reconstruction losses can be represented as
\begin{align}
    \mathcal{L}_{cyc}^{A\rightarrow B\rightarrow A}=
    \mathop{{}\mathbb{E}}\left[\lVert x_{A\rightarrow B\rightarrow A}-x_A\rVert_1\right],
    \qquad\qquad
    \mathcal{L}_{i}^{A\rightarrow A}=
    \mathop{{}\mathbb{E}}\left[\lVert x_{A\rightarrow A}-x_A\rVert_1\right].
\end{align}
We also use latent-level reconstruction losses that encourage the networks to impose style information while maintaining the original content during the forwarding phase. First, the style reconstruction loss is computed between the style features of ${(z^s_{A\rightarrow{B}}, z^s_B)}$, which makes our networks properly reflect the style because ${z^s_{A\rightarrow{B}}}$ is constrained to be equivalent to ${z^s_{B}}$. Second, the content reconstruction loss is computed between $({z^c_A, z^c_{A\rightarrow{B}})}$, and this encourages the networks to maintain the original content ${z^c_A}$ after performing a translation. These two losses can be formulated as
\begin{align}
    \mathcal{L}_{s}^{A\rightarrow B}=\mathop{{}\mathbb{E}}~[\lVert E_{B}^s(x_{A\rightarrow B})-E_B^s(x_B) \rVert_1],
    \qquad\qquad
    \mathcal{L}_{c}^{A\rightarrow B}=\mathop{{}\mathbb{E}}[\lVert E_B^c(x_{A\rightarrow B})-E_A^c(x_A) \rVert_1]
\end{align}
Lastly, the adversarial loss~\cite{goodfellow2014generative} is used to minimize the distance of the two distributions of the real images in a target domain and the generated images. For this purpose, we exploit LSGAN~\cite{mao2017least} as our adversarial loss, i.e.,
\begin{align}
    \mathcal{L}_{D_{adv}}^{B}=\tfrac{1}{2}\mathop{{}\mathbb{E}}[(D(x_B)-1)^2]+\tfrac{1}{2}\mathop{{}\mathbb{E}_{x_{A\rightarrow{B}}}}[(D(x_{A\rightarrow{B}}))^2],
    \quad
    \mathcal{L}_{G_{adv}}^{B}=\tfrac{1}{2}\mathop{{}\mathbb{E}}[(D(x_{A\rightarrow{B}})-1)^2]
\end{align}
Note that our translation backbone is trained to translate in both directions of ${(\mathcal{X}_A\rightarrow{\mathcal{X}_B}\rightarrow{\mathcal{X}_A})}$ and ${(\mathcal{X}_B\rightarrow{\mathcal{X}_A}\rightarrow{\mathcal{X}_B})}$. Finally, our full loss is formulated as
\begin{align}
    \mathcal{L}_D&=\mathcal{L}_{D_{adv}}^A+\mathcal{L}_{D_{adv}}^B\\
    \mathcal{L}_G&=\mathcal{L}_{G_{adv}}^A+\mathcal{L}_{G_{adv}}^B+\lambda_{latent}(\mathcal{L}_s+\mathcal{L}_c)+\lambda_{pixel}(\mathcal{L}_{cyc}+\mathcal{L}_i^{A\rightarrow{A}}+\mathcal{L}_i^{B\rightarrow{B}}),
\end{align}
where each term without the domain notation is bidirectionally applied within two different domains, and we empirically set ${\lambda_{latent}=1}$ and ${\lambda_{pixel}=10}$.

\begin{figure}[t]
  \includegraphics[width=\linewidth]{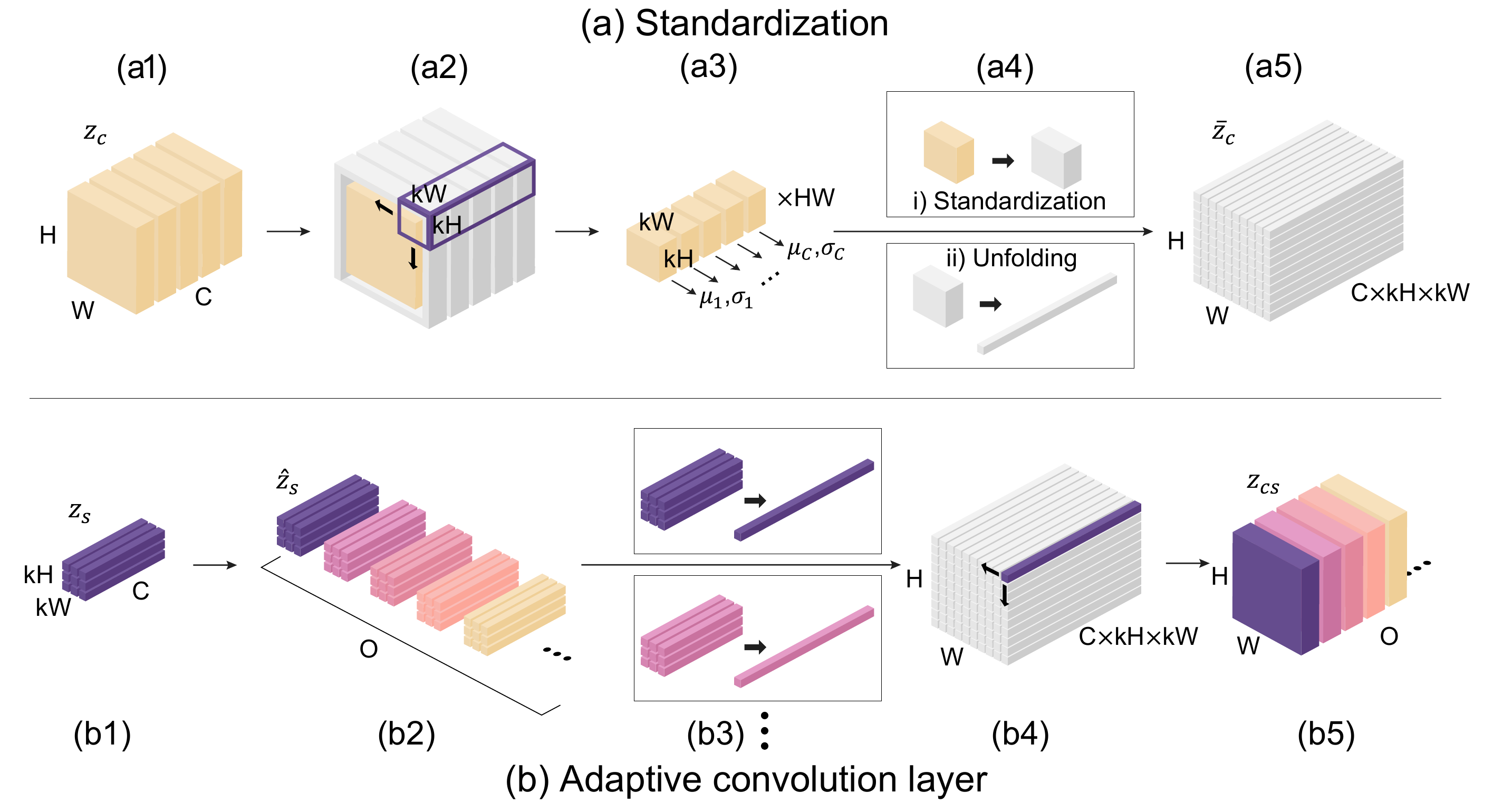}
  \vspace*{-0.4cm}
  \caption{Overview of the style branch. The first step of the style branch is (a) the local standardization step that makes each local patch of the input activation map have a zero mean and a unit variance, e.g., neutralizing the original style. The second step is (b) the style injection into the standardized local patch by applying dynamically determined convolution filters. Detailed descriptions are found in Section~\ref{subsub:style injection of AdaCoN}.}
  \label{Fig:adaptive convolution-based normalization}
  \vspace*{-0.4cm}
\end{figure}

\subsection{Adaptive convolution-based normalization (AdaCoN)}\label{sub:adaptive convolution-based normalization (AdaCoN)}
The goal of AdaCoN is to produce an output feature ${z_{out}}$ that can reflect the style of ${z_s}$ while maintaining the identity of ${z_c}$. The combined feature ${z_{out}}$ is used as input to a decoder to generate a translated image. Note that we omit the domain notation in this section for brevity. 
\subsubsection{Basic components}\label{subsub:Basic compositions of AdaCoN}
As illustrated in Fig.~\ref{Fig:Overview of the networks} (h), AdaCoN is composed of a style branch (h1) to reflect the style and a content branch (h2) that aims to maintain the content identity. Given the content ${z_c}$ and the style ${z_s}$, the style branch learns to inject the style into the content. On the other hand, the content branch learns to keep the essential information of the given ${z_c}$, so that the output of AdaCoN can maintain its original identity. Lastly, In the joining step (h3), the outputs of the branches are concatenated and forwarded into a subsequent convolution layer. Note that an additional analysis of this structure is provided in Appendix. 

\subsubsection{Style branch}\label{subsub:style injection of AdaCoN}
\paragraph{Standardization function.} ${g_{\tiny\textrm{AdaCoN}}}$ normalizes the content feature ${z_c}$ before applying adaptive convolution. Specifically, we compute the statistics of ${z_c}$ from the channel-wise local patch of the size ${kH \times kW}$, where ${kH}$ and ${kW}$ are a kernel height and a kernel width, respectively. We use ${g_{\tiny\textrm{AdaCoN}}}$ because locally computed statistics can be more effective in normalizing a given feature than globally computed ones. 
Our standardization is formulated as
\begin{align}\label{eq:adaptive convolution-based normalization, standardization}
    \Bar{z}_c=g_{\tiny\textrm{AdaCoN}}(z_c)=\frac{\phi(z_c)-\mu_{\tiny kH,kW}(\phi(z_c))}{\sigma_{\tiny kH,kW}(\phi(z_c))},
\end{align}
where ${\phi}$ denotes an unfolding operation that amasses every patch of ${z_c}$ and unites it into one tensor. Fig.~\ref{Fig:adaptive convolution-based normalization}(a) concretely describes the procedure. (a1) given ${z_c} \in \mathbb{R}^{C\times H\times W}$, (a2) ${\phi}$ extracts each sliding local block in ${\mathbb{R}^{C\times kH\times kW}}$ from the zero-padded ${z_c}$ and the extracted blocks are united into one tensor in ${\mathbb{R}^{H\times W \times C\times kH\times kW}}$. (a3) In order to perform the standardization, we compute the mean and the standard deviation along the dimensions of ${kH\times kW}$. (a4) We then normalize the content feature by exploiting its local channel-wise statistics. That is, ${g_{\tiny\textrm{AdaCoN}}}$ performs a local normalization by using statistics specified in local patch. Note that ${H}$ and ${W}$ dimensions of ${\phi(z_c)}$ imply a spatial coordinate of the local patch where it is extracted from, such that the number of patches is equivalent to ${H\times W}$. (a5) Finally, we obtain the patch-wisely normalized feature in ${\mathbb{R}^{C\cdot kH\cdot kW \times H\times W}}$.

\paragraph{Adaptive convolution layer.}
${f_{\tiny\textrm{AdaCoN}}}$ takes ${z_s}$ and ${z_c}$ as inputs and generates a stylized feature ${z_{cs}}$ as output. 
Specifically, as illustrated in Fig.~\ref{Fig:adaptive convolution-based normalization} (b1), ${f_{\tiny\textrm{AdaCoN}}}$ first takes the style feature ${z_s}\in \mathbb{R}^{C\times kH\times kW}$ as an input and (b2) encodes ${z_s}$ to the convolution weights ${\hat{z}_s} \in {\mathbb{R}^{C\times O\times kH\times kW}}$, where ${O}$ is the number of output channels. Lastly, after unfolding it to the dimensions of ${\mathbb{R}^{C\cdot kH\cdot kW \times 1\times 1}}$, we apply this weights as the form of the convolution operation and obtain the stylized feature ${z_{cs}}$ (b3-b5). Finally, the adaptive convolution is formulated as 
\begin{align}\label{eq:adaptive convolutional block}
    \hat{z}_s = \psi(z_s), \qquad
    f_{\tiny\textrm{AdaCoN}}(\Bar{z}_c, \hat{z}_s) = \sum_{k=1}^{kH}\sum_{l=1}^{kW}[\Bar{z}_c(i+k-1,j+l-1)\hat{z}_s(k,l,n)],
\end{align}
\vspace*{-0.8em}
\begin{align*}
    \;\; \qquad\qquad\qquad\qquad\qquad
    for\,i=1,...,H,\;j=1,...,W,\,and\;n=1,...,O,
\end{align*}
where ${\psi}$ represents a function that learns to properly encode a given style ${z_s}$ as the convolution weight ${\hat{z}_s}$ of ${f_{\tiny\textrm{AdaCoN}}}$. ${i}$ and ${j}$ indicates the horizontal and the vertical coordinates, respectively, and ${H}$ and ${W}$ are the height and the width of ${\Bar{z}_{c}}$, respectively. Lastly, we add the mean of the style ${\mu_s}$ to the stylized feature ${z_{cs}}$ that can be viewed as a bias in the convolution operation.

\section{Experiments}\label{sec:Experiments}

This section describes the dataset and the baseline models we used for the experiments in Section~\ref{sub:experiments implementation}. Subsequently, we discuss the comparison results with the baselines in Section~\ref{sub:Baseline comparison}. Lastly, we analyze our proposed method in detail in Section~\ref{sub:analysis}.

\subsection{Experimental settings}\label{sub:experiments implementation}

\paragraph{Dataset}
We conduct evaluations with diverse datasets. First, we use CelebA dataset~\cite{liu2015faceattributes}. This is a widely-used facial dataset involving multiple attributes. In order to construct a dataset with a large domain gap, we combine several attributes and newly form the dataset, such as (Male, Non-Bangs, Non-Smiling${\Rightarrow}$Female, Bangs, Smiling). Second, we use BAM dataset~\cite{Wilber_2017_ICCV}, composed of numerous artworks labeled with its artistic style, such as watercolor and vector-graphic. We use Watercolor $\Leftrightarrow$ Pen, Vector $\Leftrightarrow$ Pen, and Oil $\Leftrightarrow$ Pen, in order to demonstrate AdaCoN can perform image translation with a substantial domain difference. Finally, Edges $\Leftrightarrow$ Handbag~\cite{Zhu_2017} and Summer $\Leftrightarrow$ Winter~\cite{isola2017image} datasets are used to confirm the wide applicability of AdaCoN in diverse image translation tasks. We commonly set the size of the image as ${256 \times 256}$ in all the experiments.

\paragraph{Baseline methods}
We compare our proposed method with the AdaIN~\cite{huang2017arbitrary} exploited in MUNIT, and GDWCT~\cite{cho2019image}. The main difference among them lies in a specific method of combining the content feature with the style feature. As for the settings of ours, we explore various settings by adjusting the hyperparameters, such as the kernel size $\{3, 7, 11\}$ and the style dimension $\{8, 64, 128\}$ of AdaCoN. We empirically set the kernel size of 3 and the style dimension of 128. The specific results of those hyperparmeters are reported in the Section~\ref{sub:analysis}.

\paragraph{Training details}
For training the models, we exploit the Adam optimizer~\citep{kingma2014adam} with ${\beta_1=0.5}$ and ${\beta_2=0.999}$. We empirically adopt the initialization method~\citep{he2015delving} for initializing our models. We also set one for the batch size and 0.0001 for the learning rate. We regularly decay the learning rate by half in every 50,000 iteration and the decaying is started from 200,000 iterations. Every model exploited in the experiments are trained for 500,000 iterations on a NVIDIA TITAN Xp GPU for 90 hours.

\paragraph{Evaluation metric}
In order to evaluate the methods, we measure the the classification accuracy as well as content distance using a pretrained Inception-v3 model~\cite{szegedy2016rethinking}. To be specific, the content distance is measured by computing L2 distance of the features from intermediate layer of Inception-v3 between the input images and the translated ones. A lower content distance indicates that the gap between the them is relatively small. On the other hand, the evaluation on style injection is measured by the classification accuracy. This is because a well-trained image translation model can transform the domain of input image, so that a higher classification accuracy shows that the translation model successfully generates the prominent characteristics of the target domain. For training the classification model, we exploit the pretrained Inception-v3 and fine-tuned on CelebA dataset~\cite{liu2015faceattributes}. To evaluate the performance on multi-attribute translation task, we train the classifiers with multi-label dataset.

\subsection{Baseline comparison}\label{sub:Baseline comparison}
This section reports the comparison results of AdaCoN with other baseline methods. Quantitative results using the classification accuracy and the content distance are described in Section~\ref{subsub:quantitative comparison} and the qualitative results on CelebA dataset~\cite{liu2015faceattributes} is reported in Section~\ref{subsub:qualitative comparison}.

\begin{table*}[t]
\vspace*{-0.2cm}
\begin{center}
\resizebox{\textwidth}{!}{\begin{tabular}{ccccccccc}
\toprule
\textbf{Method}  
\vspace{0.1cm}
&  $\mathbf{G_1} \Rightarrow \mathbf{G_2}$
&  $\mathbf{G_2} \Rightarrow \mathbf{G_1}$
&  $\mathbf{D_1} \Rightarrow \mathbf{D_2}$
&  $\mathbf{D_2} \Rightarrow \mathbf{D_1}$
&  $\mathbf{T_1} \Rightarrow \mathbf{T_2}$
&  $\mathbf{T_2} \Rightarrow \mathbf{T_1}$
&  $\mathbf{Z_1} \Rightarrow \mathbf{Z_2}$
&  $\mathbf{Z_2} \Rightarrow \mathbf{Z_1}$ \\ [0.03cm]
\midrule
\vspace{0.1cm}
\textbf{AdaIN}  &0.173/89.5 &  0.179/88.9 & 0.162/53.8 & 0.166/58.5  &  0.196/67.5 & 0.195/51.6 &  0.192/33.8  & 0.191/82.9 \\[0.05cm]
\vspace{0.1cm}
\textbf{GDWCT}  & 0.174/90.6 &  0.190/\textbf{90.4} & 0.173/52.3 & 0.175/64.4 & 0.202/64.9 & 0.200/47.9 &  0.197/30.5  &  0.199/85.6 \\[0.05cm]
\vspace{0.1cm}
\textbf{AdaCoN}  & 0.186/\textbf{91.6} &  0.184/90.0 & 0.202/\textbf{62.3} & 0.202/\textbf{66.5} &  0.193/\textbf{67.7} & 0.197/\textbf{57.5} & 0.199/\textbf{36.5} & 0.201/\textbf{86.7} \\
\bottomrule

\end{tabular}}
\vspace{-0.1cm}
\caption{Content loss and overall classification results(\%). We bidirectionally calculate the metric with CelebA dataset~\cite{liu2015faceattributes}. Each value in the cell indicates content loss and overall classification accuracy respectively. Abbreviations: $\mathbf{G_1}$(Male), $\mathbf{G_2}$(Female), $\mathbf{D_1}$(Young, Non-Smiling), $\mathbf{D_2}$(Old, Smiling), $\mathbf{T_1}$(Non-Bald, Young, Eyeglasses), $\mathbf{T_2}$(Bald, Old, Non-Eyeglasses),  $\mathbf{Z_1}$(Male, Non-Bangs, Non-Smiling),$\mathbf{Z_2}$(Female, Bangs, Smiling)  }

\label{Table:classification_accuracy}
\vspace*{-0.3cm}
\end{center}
\end{table*}

\begin{figure}[t]
  \includegraphics[width=\linewidth]{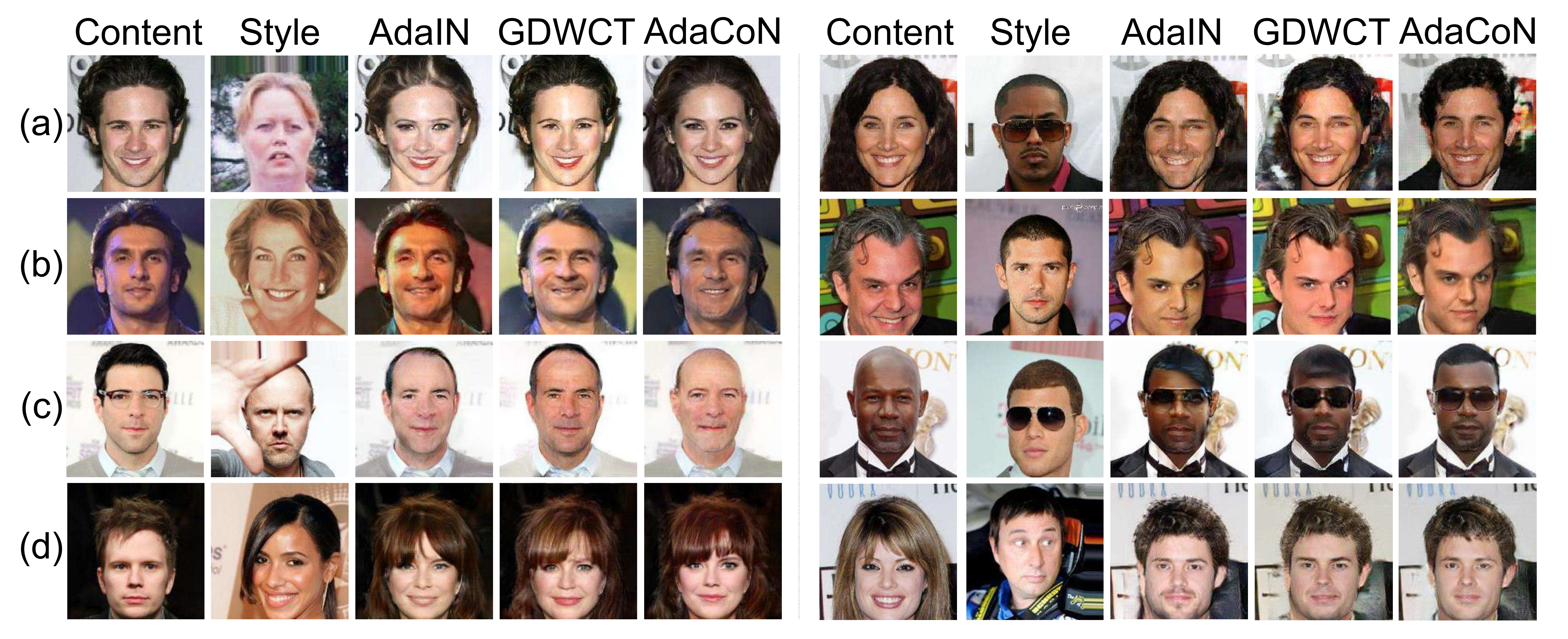}
  \vspace*{-0.7cm}
  \caption{Comparisons with baselines; (a):$\mathbf{G_1} \Leftrightarrow \mathbf{G_2}$, (b):$\mathbf{D_1} \Leftrightarrow \mathbf{D_2}$, (c):$ \mathbf{T_1} \Leftrightarrow \mathbf{T_2}$, (d):$ \mathbf{Z_1} \Leftrightarrow \mathbf{Z_1}$ }
  \label{Fig:Comparisons with baselines on CelebA dataset}
  \vspace*{-0.5cm}
\end{figure}

\subsubsection{Quantitative comparison}\label{subsub:quantitative comparison}
The classification accuracy increase when a translated output is correctly classified over every target attribute. As shown in Table.~\ref{Table:classification_accuracy}, our model displays the higher classification accuracy than other baselines. Moreover, the gap between AdaCoN and other baselines tends to be larger in multi-attribute translation task than the single attribute translation one. We believe this is because the multi-attribute translation tasks demand more considerable style injection than the single-attribute translation. For example, in case of ${(Z_1\Rightarrow Z_2)}$, in order to translate an image to the target attributes, the translation networks must change the regions of the manly characteristics, the hair, and the mouth. On the other hand, the case of ${(G_1\Rightarrow G_2)}$ requires to change only the regions of the manly characteristics, of which the amount of changes the task demands is relatively small. As for the content distance, even though AdaCoN obtains the highest score in the content distance in most translation cases, the small amount of differences ensures that AdaCoN can maintain content-identity. Considering our objective is strong reflection of the style, it is tolerable to lose the small amount of content information.

\subsubsection{Qualitative comparison}\label{subsub:qualitative comparison} 
Fig.~\ref{Fig:Comparisons with baselines on CelebA dataset} shows the comparison results of AdaCoN with baselines on various attribute translation cases. The results demonstrate that AdaCoN can significantly reflect the style compared to baselines. For example, in case of (c) in the left macro column, whose the target attributes are (Bald, Non-Eyeglasses, Old), AdaCoN considerably applies the style of the exemplar, such that the result of AdaCoN represents the bald and old man without the eyeglasses. However, both AdaIN and GDWCT keep the hair even though the style of the exemplar includes the bald attribute. On the other hand, (a) in the right macro column, of which the target attribute is Male shows the difference of the amount of the style reflection between baselines. Specifically, in order to transfer the style of man, every baseline removes the make-up. Furthermore, AdaIN makes the beard while keeping the hair length long. GDWCT incompletely removes hair region while AdaCoN clearly removes the hair region. Since the long hair is the dominant characteristic of woman, the output of AdaCoN changed to short hair verifies the superior performance of AdaCoN in style reflection. 

\subsection{Additional analysis}\label{sub:analysis}


\begin{figure}
\vspace*{0.3cm}
  \includegraphics[width=\linewidth]{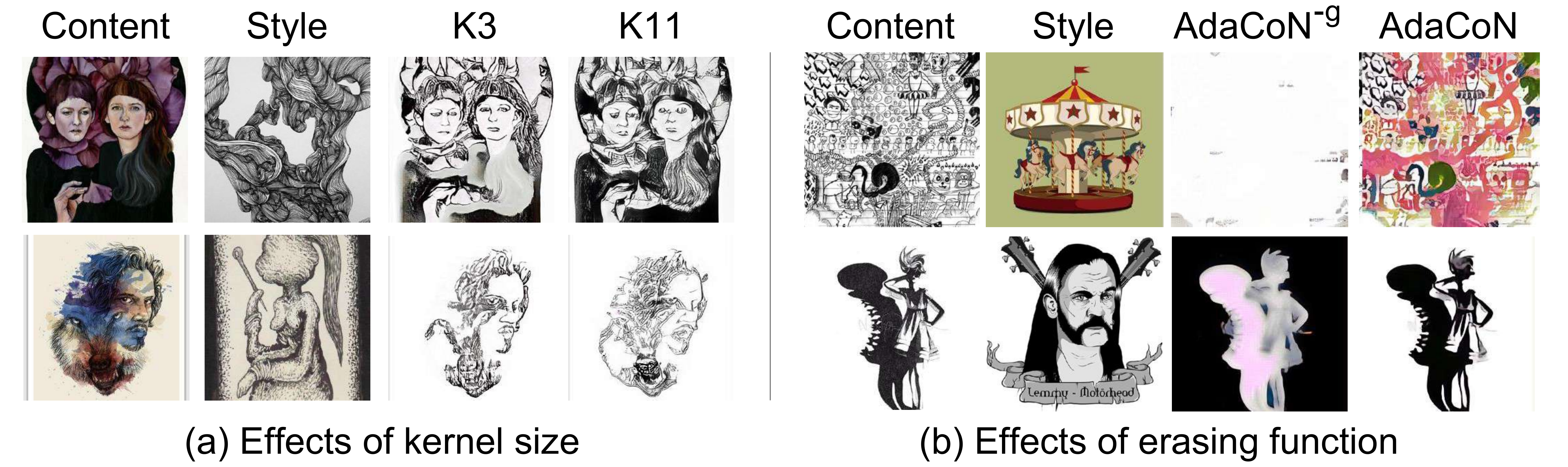}
  \vspace*{-0.3cm}
  \caption{Kernel size comparison and justification of standardization function. We perform experiments in order to explore the effects of the kernel size of AdaCoN and justify our standardization function. We exploit (Oil ${\Rightarrow}$ Pen) and (Watercolor ${\Rightarrow}$ Pen) of BAM dataset~\cite{Wilber_2017_ICCV} in (a) and (b), respectively.}
  \label{Fig:effects of kernel size and effects of standardization function}
  \vspace*{-0.2cm}
\end{figure}

\paragraph{Effects of kernel size.}
As shown in Fig.~\ref{Fig:effects of kernel size and effects of standardization function}(a), the kernel size is relevant to the spatial-awareness. In the first row, the hair color on the chest of the woman of the content image is different from the other hair color of hers. Because the small receptive field is disadvantageous to recognizing the wide hair region, K3 fails in generating the hair on the chest naturally. On the other hand, K11 shows the better results in generating the hair region because it has the larger receptive field. Furthermore, we observe that the larger kernel size engenders the larger amount of style reflection. For instance, the results of K11 more strongly reflect the style, so that it distorts the eye and mouth of the content in the first row and represents more conspicuous texture in the second row, compared to the results of K3.

\paragraph{Effects of standardization function.} Fig.~\ref{Fig:effects of kernel size and effects of standardization function}(b) shows the effects of the standardization function of AdaCoN. ${\textrm{AdaCoN}^{-g}}$ represents the results from a model trained without the standardization function ${g_{\tiny\textrm{AdaCoN}}}$. As shown in the results in both rows, ${g_{\tiny\textrm{AdaCoN}}}$ plays essential role in injecting a style because the model trained without the standardization function ${g_{\tiny\textrm{AdaCoN}}}$ fails in performing a translation. We believe this is attributed to the conflicts of the style features between the content (input) and the style (exemplar) images. Specifically, the input image has both the content and the style features, so that if its style feature is not removed by ${g_{\tiny\textrm{AdaCoN}}}$, the style feature extracted from the exemplar can give rise to the degradation of the style reflection performance. As a consequence, the results demonstrate that our proposed standardization function based on local normalization is essential in AdaCoN.

\begin{figure}
  \includegraphics[width=\linewidth]{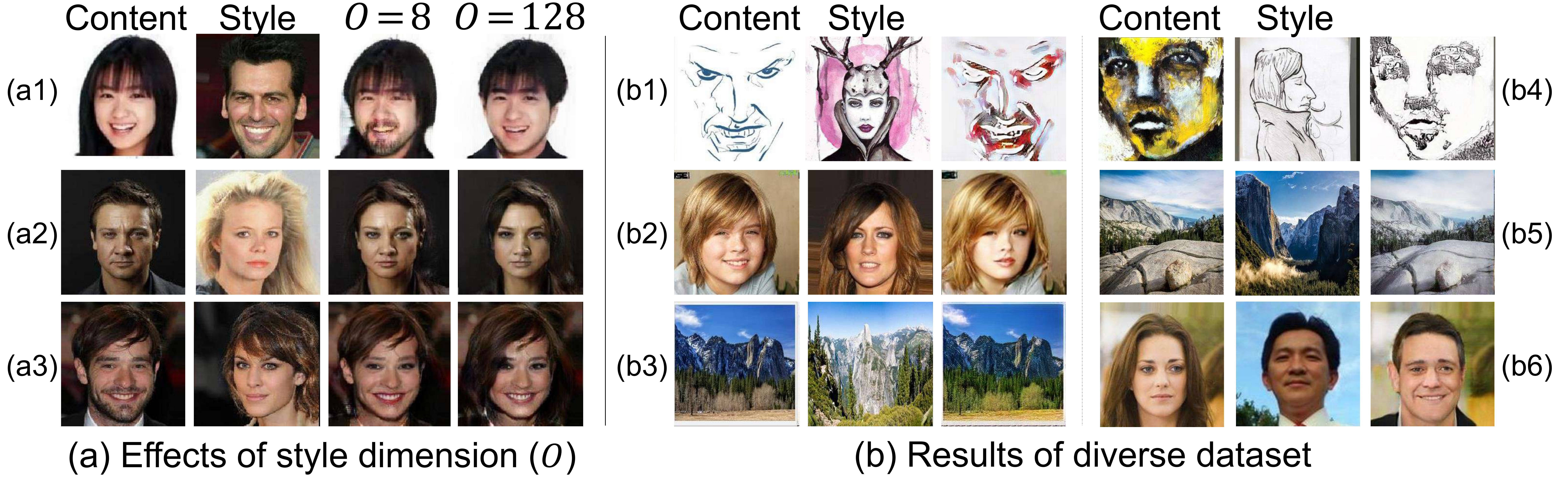}
  \vspace*{-0.4cm}
  \caption{Effects of style dimension and results from diverse dataset. (a) performs the translation of Male ${\Leftrightarrow}$ Female). (b) is conducted with (b1): Pen ${\Rightarrow}$ Watercolor, (b3): Winter ${\Rightarrow}$ Summer, (b4): Oil ${\Rightarrow}$ Pen, (b5): Summer ${\Rightarrow}$ Winter, (b2, b6): Male, Smile, Straight-Hair, Big-Nose ${\Leftrightarrow}$ Female, Non-smile, Wavy-Hair, Small-Nose, individually.}
  \label{Fig:Effects of style dimension and results of diverse dataset}
  \vspace*{-0.5cm}
\end{figure}

\paragraph{Effects of style dimension (${O}$) and results on diverse dataset.}
We compare the effects of ${O}$ that indicates the number of channels of ${z_{cs}}$. As discussed in Appendix, ${O}$ determines the extent of the style reflection to the output of AdaCoN ${z_{out}}$. As illustrated in Fig.~\ref{Fig:Effects of style dimension and results of diverse dataset}(a), the results verify that the amount of the style reflection is directly affected by ${O}$. For instance, (a1) shows the hair region of ${O=128}$ is clearly removed while ${O=8}$ relatively keeps hair region. We further observe that a beard, the other dominant characteristic of man, is rather transferred in ${O=8}$. This shows that the low dimension of ${z_{cs}}$ tends to translate the domain with the minimum change. That is, this result demonstrates that the size of ${O}$ has a positive correlation with the amount of the style reflection, such that it can be usefully exploited when attempting to control the extent of the style reflection. Meanwhile, in order to verify AdaCoN can be exploited widely as well as robustly along the diverse dataset, we conduct the experiment in Fig.~\ref{Fig:Effects of style dimension and results of diverse dataset}(b). The results consistently show that AdaCoN can translate a given image with a rich style.

\section{Conclusion}\label{sec:conclusion}
In this paper, we proposed the novel normalization method that can dramatically inject the style of the given exemplar in a image translation. AdaCoN locally performs the standardization of the content representation in order to properly reflect the given style, and the adaptive convolution layer, whose weights are dynamically extracted from the style encoding is applied to the standardized feature. We verify the superior performance of AdaCoN in drastic style injection through the experiments. We believe AdaCoN can be usefully exploited in diverse challenging image translation tasks that have a large gap between a source and a target domain, such as the multi-attribute translation. Finally, AdaCoN can be potentially used by incorporating an additional information with our novel normalization technique in various tasks such as object detection and semantic segmentation.



\clearpage

{\small
\bibliographystyle{unsrt}
\bibliography{mybib}
}

\clearpage

\title{Supplementary material}
\maketitle
\author{}
\date{}

\setcounter{section}{5}

\section{Appendix}

\subsection{Analysis on existing methods}\label{sub:Analysis on Combining Methods}
In order to intensively comprehend the existing methods, this section reviews their principal operations and performs the comparative analysis of them.

\subsubsection{Review on baselines}\label{subsub:review on baseline methods}
The previous methods are typically composed of two steps, of which the first step is to normalize the content feature, and the second step is to reflect the style feature to the normalized content. We formulate this procedure as ${f(g(c), s)}$, where ${g}$ and ${f}$ represent the standardization and the style injection function, respectively. In this point of view, AdaIN can be illustrated as
    \begin{align}\label{eq:AdaIN}
        g_{\tiny\textrm{AdaIN}}(c)=\frac{c-\mu_{\tiny H,W}(c)}{\sigma_{\tiny H,W}(c)}, && f_{\tiny\textrm{AdaIN}}(g(c), s) = \sigma_{\tiny H,W}(s)(g(c))+\mu_{\tiny H,W}(s),
    \end{align}
where ${H}$ and ${W}$ are the height and the width of an input feature. Each channel is normalized and combined independently. ${\sigma_{\tiny H,W}}$ and ${\mu_{\tiny H,W}}$ respectively denote the standard deviation and the mean computed along the ${H}$ and ${W}$ dimensions. In Eq.~\eqref{eq:AdaIN}, the function ${g}$ normalizes an input content feature with the channel-wise mean and variance. On the other hand, the function ${f}$ transfers the mean ${\mu_{\tiny H,W}}(s)$ and the variance ${\sigma_{\tiny H,W}}(s)$ of the style to those of the normalized content ${g(c)}$. Meanwhile, GDWCT can be represented as
    \begin{align}\label{eq:GDWCT}
        g_{\tiny\textrm{GDWCT}}(c)=Q_c\Lambda_c^{-\frac{1}{2}}Q_c^T(c-\mu_{\tiny H,W}(c)), && f_{\tiny\textrm{GDWCT}}(g(c), s)=Q_s\Lambda_s^{\frac{1}{2}}Q_s^Tg(c)+\mu_{\tiny H,W}(s),
    \end{align}
where the matrices ${\{Q_c\Lambda_c Q^T_c, Q_s\Lambda_s Q^T_s\}}$ can be obtained by the eigendecomposition of the channel covariance matrix of the content and the style features, respectively. Each of ${\{Q_c,Q_s\}}$ indicates a square matrix composed of the eigenvectors, and ${\{\Lambda_c,\Lambda_s\}}$ are diagonal matrices whose each diagonal entry indicates an eigenvalue of a corresponding eigenvector in ${\{Q_c,Q_s\}}$. In Eq.~\eqref{eq:GDWCT}, the function ${g}$ plays a similar role to Eq.~\eqref{eq:AdaIN}, but forces the more strict rule, so it normalizes not only the mean and the variance but also the covariance of an input feature by making its covariance matrix the identity matrix. As for the style injection function ${f_{\tiny\textrm{GDWCT}}}$, it matches the first and the second-order statistics of normalized content feature to those of the style feature. 
\subsubsection{Comparative analysis on baselines}\label{subsub:Comparative analysis on baselines}
The differences of the existing methods are clear when we regard those methods as a special case of the convolution operation. ${f_{\tiny\textrm{AdaIN}}}$ in Eq.~\eqref{eq:AdaIN} can be represented as the $1 \times 1$ depth-wise convolution with the bias since adaptive parameters of ${f_{\tiny\textrm{GDWCT}}}$ identically scale and shift along channels. Meanwhile, ${f_{\tiny\textrm{GDWCT}}}$ in Eq.~\eqref{eq:GDWCT} can be viewed as the $1 \times 1$ convolution layer, of which the weights are ${Q_s\Lambda^{\frac{1}{2}_s} Q^T_s}$ and the bias is ${\mu_{\tiny H,W}(s)}$. This is because the vector-matrix multiplication of a row vector of ${Q_s\Lambda^{\frac{1}{2}_s} Q^T_s} \in  {\mathbb{R}^{C\times C}}$ by the matrix ${g(c)} \in {\mathbb{R}^{C\times HW}}$ generates a new row vector in ${\mathbb{R}^{1\times HW}}$. This is identical to the $1 \times 1$ convolution operation, whose the output channel is one. From the aforementioned view, we can intensively explore these style injection functions. ${f_{\tiny\textrm{AdaIN}}}$ can be expected to transfer the lowest amount of style as it injects the style along channel, such that it engenders a relatively high consistency with the content compared to other methods. On the other hand, ${f_{\tiny\textrm{GDWCT}}}$ can be thought as a stronger combining method than ${f_{\tiny\textrm{AdaIN}}}$ because it generates the channel dimension of the content feature as a linear combination of the content feature channels. Even though GDWCT accomplishes more drastic changes of the style compared to AdaIN since it carries out  mixing channel information of the content, we claim that even more dramatic changes can be achieved if the spatial information is simultaneously considered. Hence, we propose ${n\times n}$ adaptive convolution-based normalization, whose weights are extracted from the style. We believe this can increase a transferring capacity of a given style.

\subsection{Discussion on branch-separation}\label{sub:discussion on branch-separation}
Fully exploiting the adaptive convolution-based normalization at the intermediate layers may engender considerable distortions of the content information because the spatial information as well as the channel information of the output features of AdaCoN is entirely different from those of the input features. Considering one of the task objectives is maintaining an input identity, we posit that a combination of the adaptive convolution-based normalization with the general convolution layer is reasonable choice for performing the translation. Moreover, through separating branches, we can control the amount of the style injection by changing ${O}$ that indicates the number of style dimensions. That is, the small ${O}$ gives rise to the low injection of the style.

\subsection{Additional results}
\begin{figure}
  \includegraphics[width=\linewidth]{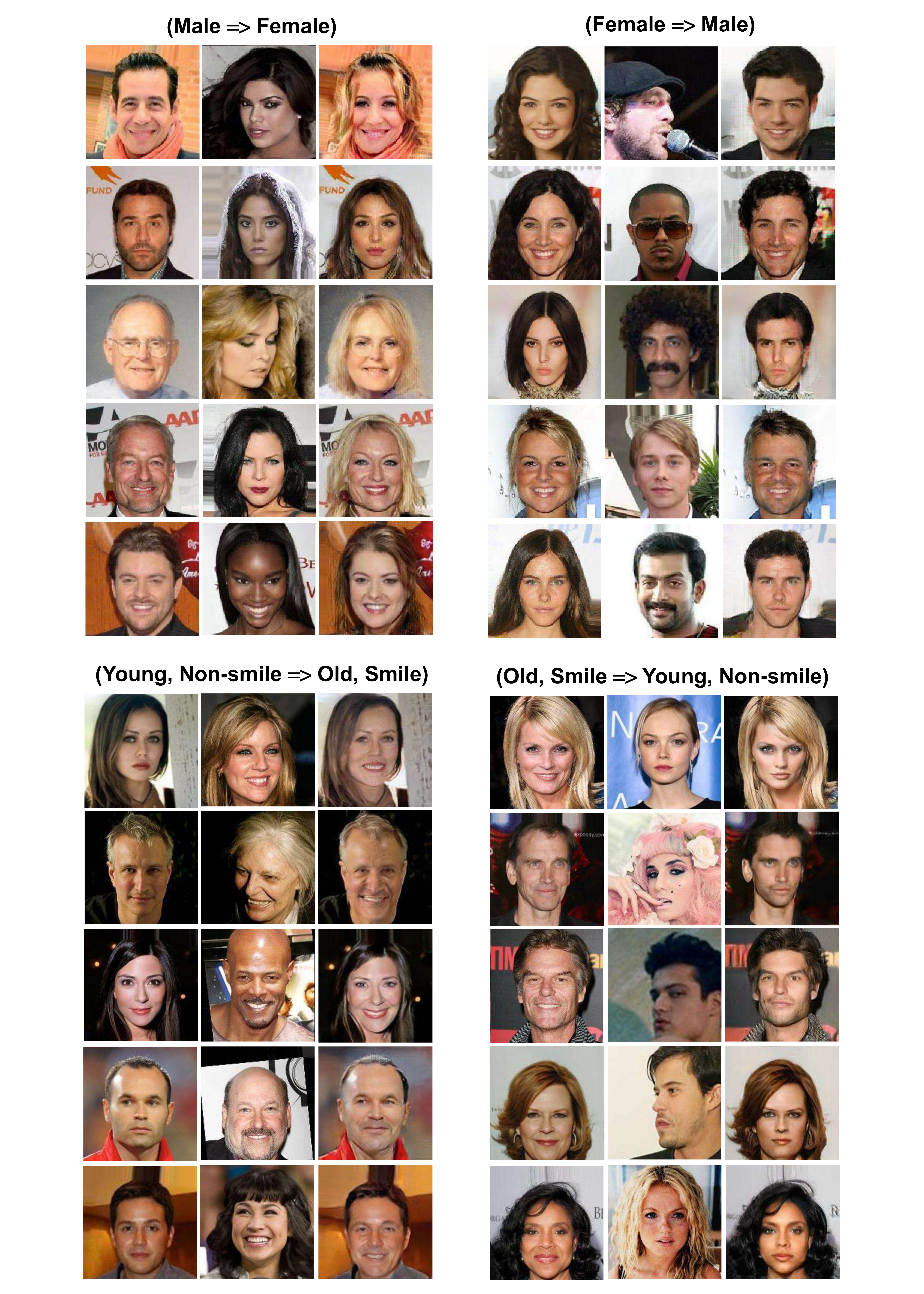}
  \caption{Extra results of our model on CelebA dataset.}
  \label{Fig:additional results1}
\end{figure}

\begin{figure}
  \includegraphics[width=\linewidth]{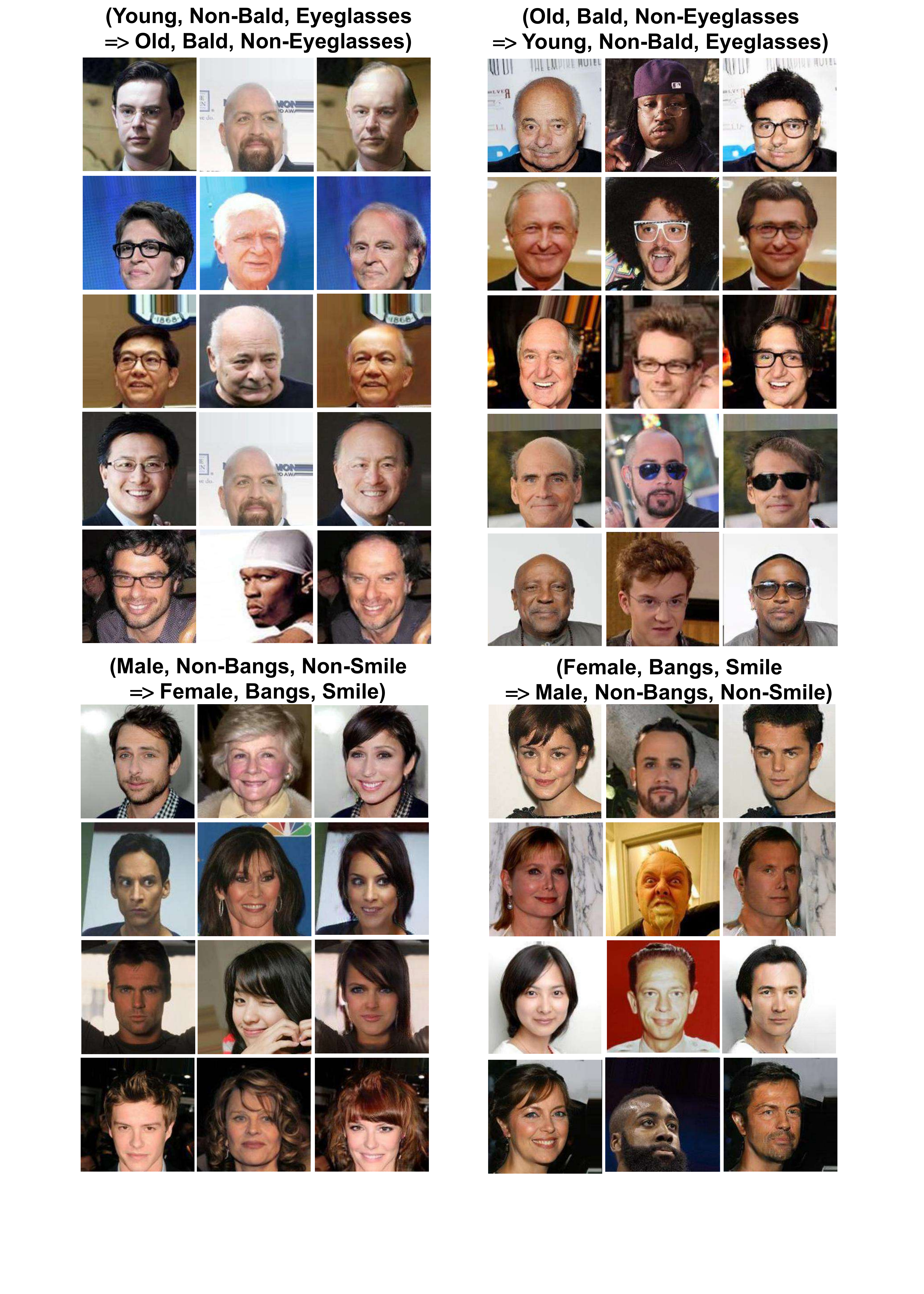}
  \vspace*{-2.1cm}
  \caption{Extra results of our model on CelebA dataset.}
  \label{Fig:additional results2}
\end{figure}

Fig.~\ref{Fig:additional results1},~\ref{Fig:additional results2} show the additional results of AdaCoN on the various image translation tasks.


\clearpage


\end{document}